\documentclass[sigconf,nonacm]{acmart}

\usepackage{tabularx}
\usepackage[utf8]{inputenc} %
\usepackage[T1]{fontenc}    %
\usepackage{url}            %
\usepackage{amsfonts}       %
\usepackage{lipsum}
\usepackage{fancyhdr}       %
\usepackage{graphicx}%
\graphicspath{{media/}}     %
\usepackage{etoolbox}       
\usepackage{makecell}  
\usepackage{soul}
\usepackage{url}
\usepackage[utf8]{inputenc}
\usepackage{caption}
\usepackage{amsmath}
\usepackage{amssymb}
\usepackage{pifont}
\usepackage{wasysym}
\usepackage{booktabs}
\usepackage{algorithm}
\usepackage{algorithmic}
\usepackage{enumitem}
\usepackage{wrapfig}
\usepackage{adjustbox}
\usepackage{xspace}
\usepackage{listings}
\usepackage{subcaption}
\usepackage[most]{tcolorbox}

\usepackage{multicol}
\usepackage{multirow}
\usepackage{bbding}
\usepackage{circledsteps}

\newtcolorbox{promptbox}{
    colback=gray!5,          
    colframe=gray!70!black,  
    coltitle=white,          
    title=Case study - CoT Prompts example, 
    fonttitle=\bfseries,     
    colbacktitle=gray!70!black, 
    boxrule=0.75pt,          
    arc=3pt,                 
    left=6pt,                
    right=6pt,               
    top=5pt,                 
    bottom=5pt,              
    boxsep=2pt,              
    fontupper=\sffamily\small  
}

\definecolor{deepred}{HTML}{CC3333}
\usepackage{amsfonts,amssymb}
\usepackage{listings}
\definecolor{myblue}{HTML}{3399CC}
\definecolor{myred}{HTML}{993333}
\definecolor{boxblue}{HTML}{248f6b}
\definecolor{lightpink}{RGB}{255, 230, 230} 
\definecolor{darkred}{RGB}{185, 38, 74}      
\usepackage{mdframed}
\usepackage{tcolorbox}
\usepackage{enumitem}
\usepackage{algorithm}
\usepackage{algorithmic}

\usepackage[most]{tcolorbox}

\AtBeginDocument{%
  }
\newcommand{\methodFont}{\texttt}
\usepackage{xspace}

\newcommand{\ours}{\methodFont{MD-UQ}\xspace}
\setcopyright{acmlicensed}
\copyrightyear{2025}
\usepackage{caption}
\usepackage{multirow}
\usepackage[capitalize]{cleveref}

\begin{document}

\title{Uncertainty Quantification of Large Language Models through Multi-Dimensional Responses}


\author{Tiejin Chen}
\affiliation{%
  \institution{Arizona State University}
  \city{Tempe}
  \country{AZ}}
\author{Xiaoou Liu}
\affiliation{%
  \institution{Arizona State University}
  \city{Tempe}
  \country{AZ}}

\author{Longchao Da}
\affiliation{%
  \institution{Arizona State University}
  \city{Tempe}
  \country{AZ}}

\author{Jia Chen}
\affiliation{%
  \institution{University of California, Riverside}
  \city{Riverside}
  \country{CA}}

\author{Vagelis Papalexakis}
\affiliation{%
  \institution{University of California, Riverside}
  \city{Riverside}
  \country{CA}}

\author{Hua Wei}
\affiliation{%
  \institution{Arizona State University}
  \city{Tempe}
  \country{AZ}}

\begin{abstract}
Large Language Models (LLMs) have demonstrated remarkable capabilities across various tasks due to large training datasets and powerful transformer architecture. However, the reliability of responses from LLMs remains a question. Uncertainty quantification (UQ) of LLMs is crucial for ensuring their reliability, especially in areas such as healthcare, finance, and decision-making.  Existing UQ methods primarily focus on semantic similarity, overlooking the deeper knowledge dimensions embedded in responses. We introduce a multi-dimensional UQ framework that integrates semantic and knowledge-aware similarity analysis. By generating multiple responses and leveraging auxiliary LLMs to extract implicit knowledge, we construct separate similarity matrices and apply tensor decomposition to derive a comprehensive uncertainty representation. This approach disentangles overlapping information from both semantic and knowledge dimensions, capturing both semantic variations and factual consistency, leading to more accurate UQ. Our empirical evaluations demonstrate that our method outperforms existing techniques in identifying uncertain responses, offering a more robust framework for enhancing LLM reliability in high-stakes applications.

\end{abstract}




\maketitle

\section{Introduction}

Large Language Models (LLMs) have demonstrated exceptional capabilities in tasks ranging from code generation to clinical decision support \citep{du2024evaluating,cascella2023evaluating}. However, their deployment in high-stakes domains requires rigorous reliability verification, as studies reveal persistent hallucination and overconfidence issues \citep{huang2024trustllm,yao2023llm}. This challenge intensifies with commercial LLMs like GPT-4, Claude, and Gemini \citep{achiam2023gpt-4,anthropic2024claude, team2023gemini}, where black-box access to architectures and parameters prevents internal confidence measurement.

While uncertainty quantification (UQ) methods form the foundation of reliable AI systems \citep{wang2025aleatoric}, existing approaches face fundamental limitations in natural language generation (NLG). Traditional UQ techniques for classification \citep{gal2016dropout} or regression \citep{ye2024uncertainty} fail to address NLG's unique challenges: semantic equivalence despite lexical variation \citep{kuhn2023semantic}, and knowledge coherence beyond surface patterns \citep{choi2024fact}. Current LLM UQ methods exacerbate these issues by focusing on single uncertainty dimensions—either semantic entropy requiring white-box access \citep{kuhn2023semantic} or knowledge-augmented approaches that propagate redundancy \citep{da2024llm}.

\begin{figure}[h]
    \centering
    \includegraphics[width=0.48\textwidth]{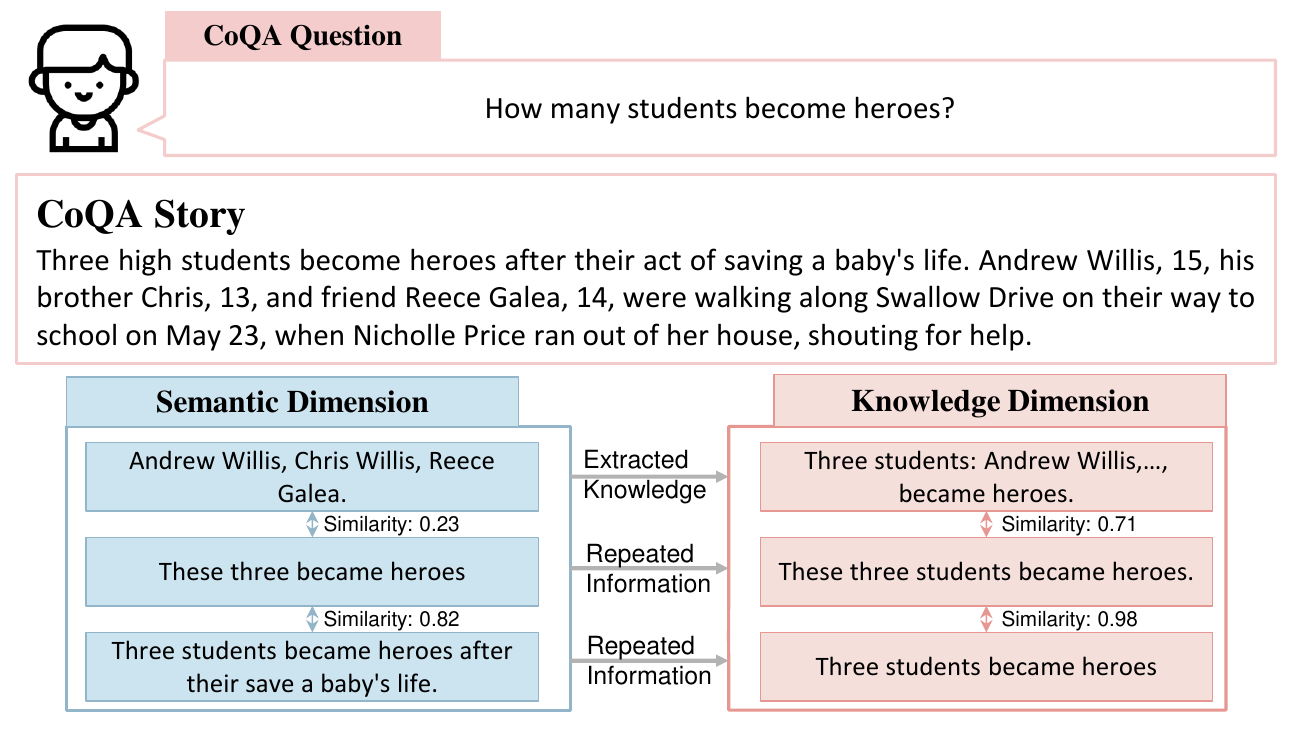}
    \caption{An example from CoQA illustrates that responses based on the knowledge dimension can extract implicit information from semantic responses while still preserving common data. Therefore, an entangled relationship between semantic and knowledge dimensions is necessary.}
    \vspace{-5mm}
    \label{fig:Knowledge_example}
\end{figure}

Effective uncertainty quantification requires analyzing complementary semantic and knowledge dimensions: (1) Semantic equivalence captures meaning consistency across paraphrased responses, while (2) Knowledge consistency verifies factual alignment with external evidence. Single-dimensional approaches prove inadequate—semantic entropy fails to detect knowledge contradictions in meaning-preserving responses \citep{da2024llm}, while knowledge-augmented methods cannot distinguish factual redundancy from semantic novelty \citep{hu2024decomposition}. This multi-dimensional nature of LLM uncertainty necessitates joint analysis of semantic and knowledge coherence features.
\begin{figure}[h]
    \centering
    \includegraphics[width=0.48\textwidth]{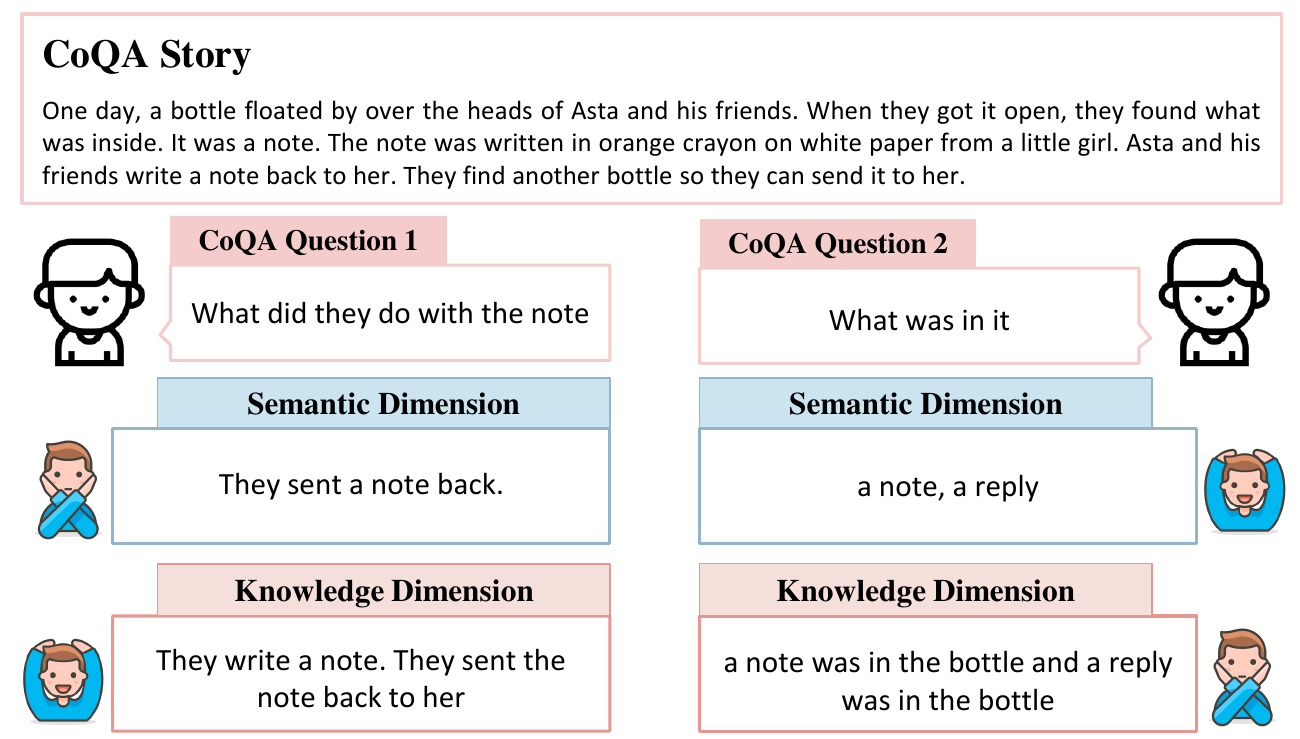}
    \caption{An example from CoQA illustrates the dynamic relationship between semantic dimension and knowledge dimension. The example shows that we need to consider both dimensions of responses.}
    \vspace{-7mm}
    \label{fig:dynamic_example}
\end{figure}
The development of such multi-dimensional UQ faces two fundamental challenges:
(1) \textbf{Semantic-Knowledge Entanglement}: Current methods conflate semantic equivalence with knowledge consistency, failing to distinguish between meaning-preserving variations and factual contradictions. We show an example in \cref{fig:Knowledge_example}. To the question \textit{How many students become heroes?}, two responses \textit{Andrew Willis, Chris Willis, Reece
Galea} and \textit{These three became heroes} differ significantly in their semantic representation, as one lists specific names while the other generalizes the information. However, they convey the same underlying knowledge: three persons became heroes.  On the other hand, the response \textit{These three became heroes} and the response \textit{Three students became heroes} exhibit similarity in both semantic and knowledge dimensions. Therefore, it is crucial to disentangle these two dimensions.
(2) \textbf{Dynamic Relationship}: The relative importance of semantic and knowledge dimensions varies across conversational contexts.  We show an example from CoQA in \cref{fig:dynamic_example}. In CoQA, questions like \textit{"What was in it"} rely heavily on semantic equivalence to ensure coherence with prior turns and knowledge extraction might misunderstand the response \textit{a note, a reply} to there are two things in it. In contrast, questions such as \textit{"What did they do with the note"} demand knowledge consistency to verify alignment with evidence in the passage. Therefore, Static fusion strategies fail to adapt to such dynamic shifts, leading to suboptimal uncertainty estimates.

To address these challenges, We propose \ours, a \textbf{M}ulti-\textbf{D}imensional \textbf{U}ncertainty \textbf{Q}uantification framework that integrates multiple dimensions of uncertainty and adaptively fuses them into a unified measure. Specifically, to integrate and disentangle information from multiple uncertainty dimensions, our method first concatenates similarity matrices from each dimension and then applies tensor decomposition to generate reconstruction errors through orthogonal analysis. Additionally, to account for dynamic inter-dimensional relationships, we adaptively ensemble reconstruction errors obtained from different tensor decomposition methods, ensuring robust and accurate uncertainty estimation across diverse tasks. Extensive experiments demonstrate the effectiveness of multi-dimensional analysis and show that our framework achieves superior performance, particularly on challenging datasets. In summary, the contributions of this paper are:

\begin{itemize}
    \item We are the first to propose an uncertainty quantification framework that integrates multiple dimensions of uncertainty, including semantic and knowledge coherence. Our method captures complementary uncertainty signals by combining information from implicit knowledge and original responses, enabling a more comprehensive analysis of black-box LLM outputs.
    \item We develop a novel tensor decomposition approach to orthogonally analyze multi-dimensional data and represent uncertainty using reconstruction errors. This process separates repeated information, reduces estimation bias, and disentangles semantic and knowledge features effectively.
    \item Extensive experimental results on CoQA~\cite{reddy2019coqa}, NQ\_Open~\cite{kwiatkowski2019natural} and HotpotQA~\cite{yang2018hotpotqa} datasets show that compared with state-of-the-art methods, our framework consistently performs better under various settings.
    
\end{itemize}

\section{Preliminaries}
\label{sec:background}

\subsection{Problem Formulation}
\label{subsec:problem}
Modern uncertainty quantification (UQ) for black-box LLMs operates through two sequential stages: similarity measurement between responses and uncertainty estimation from these similarities. Let $\mathcal{M}$ denote a black-box LLM generating $n$ responses $\{A^1,\ldots,A^n\}$ to input $Q$. The UQ task estimates confidence $U $ through:

\begin{equation}
U = f(\mathbf{S}), \quad \mathbf{S} \in \mathbb{R}^{n\times n}
\end{equation}

\noindent where $\mathbf{S}$ is the similarity matrix with the $(i, j)$-th entry capturing the proximity  of responses $A^i$ and $A^j$, and $f$ represents the estimation strategy. This formulation enables UQ without accessing internal model probabilities.

\subsection{Measuring Response Similarities}
\label{subsec:similarities}
The foundation of reliable UQ lies in effective similarity measurement. We analyze two complementary approaches capturing different aspects of response quality.

\subsubsection{Semantic Dimension}
\label{subsubsec:semantic}
Semantic similarity focuses on surface-level consistency between responses. The Jaccard index \citep{lin2023generating} offers simple lexical comparison:

\begin{equation}
s_{ij} = \frac{|A^i \cap A^j|}{|A^i \cup A^j|}
\end{equation}

While computationally efficient, Jaccard ignores word order and semantics. For deeper analysis, there is another common way to compute the semantic similarity, which uses DeBERTa's NLI capabilities~\citep{he2021deberta}:

\begin{equation}
s_{ij} = \frac{1}{2}\left(P_{\text{entail}}(A^i,A^j) + P_{\text{entail}}(A^j,A^i)\right)
\end{equation}

Where $P_{\text{entail}}(A^i,A^j)$ the probability of $A^i$ entails $A^j$ that is output from the NLI model.

Compared with the Jaccard index, NLI-based scoring better captures semantic equivalence but remains sensitive to syntactic variation. For instance, paraphrased factual statements may receive low scores despite equivalent meaning. To the question \textit{How many students became heroes}, The response \textit{These three became heroes} and the response \textit{Andrew Willis, Chris Willis, Reece Galea} share the factual knowledge that three students became heroes while their similarity from NLI models will be low as 0.015. Therefore, relying solely on responses from the semantic dimension may result in information loss.

\subsubsection{Knowledge Dimension} 
\label{subsubsec:knowledge}

The knowledge dimension operates through a structured pipeline that transforms raw responses into factual representations. 
Given a question $Q$ and its original response $A^i$, a knowledge representation $K^i$ could be generated through a knowledge mapping process by extracting explicit claims: $K^i = \mathcal{M}_{\text{aux}}(Q, A^i)$ and augmenting the response, where  $\mathcal{M}_{\text{aux}}$ denotes for an auxiliary LLM. Specifically, we could use an LLM to augment with prompts taking into the question and original response:

\begin{tcolorbox}[colback=gray!5!white, colframe=gray!75!green, title=\textbf{Prompt Example for Knowledge Mapping}] 

 \ding{182}: Extract all factual claims from this response $\langle A^i \rangle$, phrased as standalone statements independent of specific wording. 

 \ding{183}: Include only information directly relevant to answering the question: $\langle Q \rangle$.

\end{tcolorbox}

This claim extraction disentangles implicit knowledge from surface semantics and removes stylistic variations while preserving core factual content.

\subsection{Estimating Uncertainty}
\label{subsec:estimation}
With similarity matrices constructed, existing UQ methods employ two principal ways to estimate uncertainty, each offering unique advantages and limitations.

\subsubsection{Number of Semantic Sets (UNumSet)}
\label{subsubsec:numset}
Proposed by \citet{kuhn2023semantic}, this method groups responses into equivalence classes using bidirectional entailment checks from an NLI model. Formally, responses $A^i$ and $A^j$ are merged into the same semantic set if:
\begin{equation}
P_{\text{entail}}(A^i, A^j) > P_{\text{contra}}(A^i, A^j) \quad \text{and} \quad P_{\text{entail}}(A^j, A^i) > P_{\text{contra}}(A^j, A^i).
\end{equation}
The uncertainty measure $U_{\text{NumSet}}$ equals the number of resulting semantic sets. This approach aligns with spectral graph theory. Because when using binary adjacency matrices ($W_{ij} \in {0,1}$), the number of zero eigenvalues in the graph Laplacian corresponds to the number of connected components \citep{von2007tutorial}. While this method discretizes continuous semantic relationships, it fails to capture partial meaning overlaps.

\subsubsection{Graph Laplacian}
\label{subsubsec:laplacian}
Building on spectral graph principles \citep{agaskar2013spectral,lin2023generating}, this method quantifies uncertainty through the eigenvalues ${\lambda_k}$ of the normalized graph Laplacian $L = I - D^{-1/2}WD^{-1/2}$:
\begin{equation}
U_{\text{EigV}} = \sum_{k=1}^n \max(0, 1 - \lambda_k).
\end{equation}

Here, eigenvalues $\lambda_k$ encode connectivity. fragmented graphs (low consistency in responses and thus high uncertainty) have more small eigenvalues. Compared with $U_{\text{NumSet}}$, this method is able to capture possible overlapping semantic relationships.

\subsection{Dimensional Analysis}
Now, we compare the difference between similarity matrices from knowledge and semantic dimensions. In detail, we use the NLI model to obtain the similarity matrix. In \cref{tab:similarity_matrix_stat}, we present the mean values and the proportion of similarity scores greater than 0.55 in the similarity matrices. The results show that similarity matrices in the knowledge dimension have more large value as well as a larger mean value. This reveals the knowledge dimensions' superior consistency. These results highlight the importance of multi-dimensional analysis since semantic features capture response variability, while knowledge features track factual consistency hidden behind the semantic features. Our tensor decomposition effectively combines these complementary signals.

\begin{table}[t!]
    \centering
    \begin{tabular}{lcc}
        \toprule
        Dataset & Proportion (\%) & Mean Value \\
        \midrule
        \multicolumn{3}{c}{\textbf{Semantic Similarity}} \\
        \midrule
        Coqa & 52.97 & 0.5430 \\
        nq\_open & 17.11 & 0.1839 \\
        Trivialqa & 51.15 & 0.5154 \\
        \midrule
        \multicolumn{3}{c}{\textbf{Knowledge Similarity}} \\
        \midrule
        Coqa & 57.40 & 0.5723 \\
        nq\_open & 31.16 & 0.3281 \\
        Trivialqa & 60.17 & 0.6058 \\
        \bottomrule
    \end{tabular}
    \caption{Proportion of similarity values greater than 0.55 and mean similarity values for the similarity matrices in the semantic space and the knowledge space. The results show that the knowledge similarity matrix has larger values.}
    \vspace{-10mm}
    \label{tab:similarity_matrix_stat}
\end{table}
\section{Methodology}
In this section, we introduce our methods: Multi-Dimensional Uncertainty Quantification~(MD-UQ).

\begin{figure*}[h]
    \centering
    \includegraphics[trim=0 360 0 10, clip,width=\textwidth]{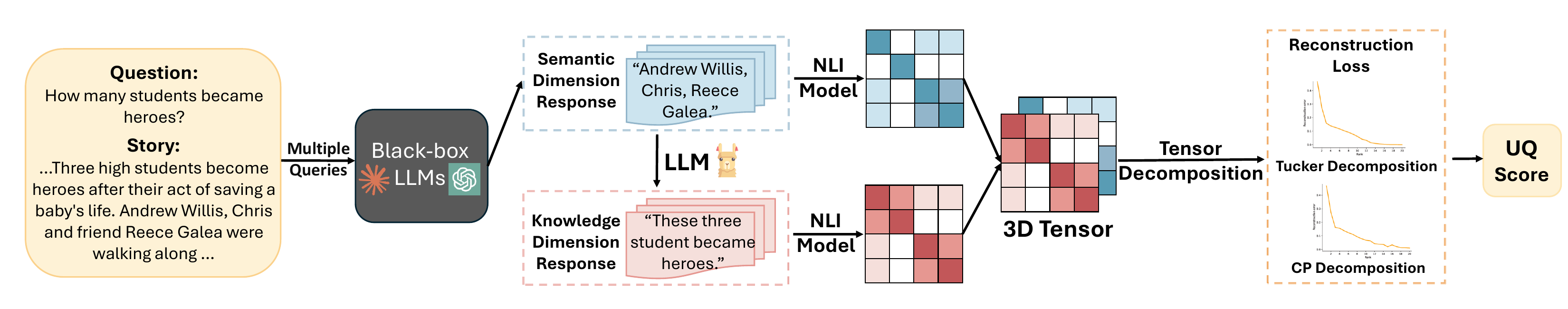}
    \caption{The overall pipeline of \ours. \ours utilizes tensor decomposition methods to ensure the information from both semantic and knowledge dimension responses could be fully utilized. In the example comparison of semantic dimension and knowledge dimension, we found that though knowledge dimension responses contain implicit knowledge and revise the unreasonable similarity, both dimension responses still share a lot of common information.}
    \vspace{-3mm}
    \label{fig:pipeline}
\end{figure*}

\subsection{Overview}

While our methodological framework generalizes to multiple response dimensions, this work focuses on two well-established and complementary dimensions: the \textit{semantic dimension} capturing surface-level linguistic patterns, and the \textit{knowledge dimension} encoding factual consistency \citep{da2024llm,choi2024fact,vashurin2024benchmarking}. 
The semantic similarity matrix $\mathbf{S}$ is computed using established natural language inference techniques \citep{he2021deberta}, while $\mathbf{S}_K$ leverages auxiliary LLM-generated claim comparisons as detailed in \cref{subsubsec:knowledge}. By jointly analyzing these dimensions through tensor operations, our framework addresses the limitations of single-axis approaches documented in \cref{sec:background}, particularly the inability to distinguish lexical variation from factual inconsistency.
As illustrated in \cref{fig:pipeline}, our approach integrates these dimensions through three key phases: 

\begin{enumerate}
\item \textbf{Tensor Representation} (\cref{subsec:tensor_rep}): Construct a multi-dimensional similarity tensor $\mathcal{S} \in \mathbb{R}^{n\times n\times 2}$ by concatenating semantic ($\mathbf{S}$) and knowledge ($\mathbf{S}K$) similarity matrices derived from $n$ responses
\item \textbf{Tensor Decomposition} (\cref{subsec:decomp}): Apply orthogonal tensor decomposition to isolate dimension-specific features and suppress redundant information
\item \textbf{Ensemble Scoring} (\cref{subsec:ensemble}): Combine decomposition residuals across dimensions to compute the final uncertainty measure $U_{\text{final}}$
\end{enumerate}

\subsection{Tensor Representation}
\label{subsec:tensor_rep}

After obtaining the similarity matrices $\mathbf{S}$ and $\mathbf{S}_K$, the next question will be how to fully utilize two similarity matrices to represent the information from multiple dimensions. There are many methods such as concatenating two similarity matrices within 2D dimensions:
\begin{equation}
    S_{\text{concat}} = \begin{bmatrix} \mathbf{S} \\ \mathbf{S}_K \end{bmatrix} \in \mathbb{R}^{n\times (n+n)},
\end{equation}
and then performing singular value decomposition. Therefore, we need to choose a method based on the feature of two similarity matrices. We find that though the knowledge dimension of responses could contain the implicit knowledge, $\mathbf{S}_K$ and $\mathbf{S}$ still share large repeated information. Since we actually care about the structure of both matrices, we focus on the spectral norm of $\mathbf{S}$ and $\mathbf{S}_K$ because the spectral norm captures the most dominant features of their structures. Therefore we use spectral norm ratio (SNR)~\cite{wang2020theoretical} to quantify the redundancy between the matrices:
\begin{equation}
   SNR = \frac{\|\Lambda_{\text{sem}}\|_2}{\|\Lambda_{\text{know}}\|_2},
\end{equation}

where $\Lambda_{\text{sem}}$ and $\Lambda_{\text{know}}$ denote the largest eigenvalue (i.e. the spectral norm) of $\mathbf{S}$ and $\mathbf{S_k}$, respectively. A SNR close to 1 indicates that $\mathbf{S}$ and $\mathbf{S_k}$
share a similar structure, implying a higher degree of repeated information. We calculate the mean SNR for every $\mathbf{S}$ and $\mathbf{S_k}$ in Coqa is 0.89, which demonstrates a high degree of repeated information between the two dimensions. 

For such large repeated information, if we consider using $S_{concat}$, the repeated information will dominate and thus lead to a sub-optimal result. To address the repeated information, we applied tensor decomposition, which can analyze the information from two matrices orthogonally. 
In detail, we construct a third-order tensor \( \mathcal{S} \) by stacking the semantic similarity matrix \( \mathbf{S} \) and the knowledge similarity matrix \( \mathbf{S}_K \) along a new dimension. We use $\mathcal{S} \in \mathbb{R}^{n\times n \times 2}$ to denote the concatencated tensor.

\subsection{Tensor Decomposition}
\label{subsec:decomp}
In our work, we utilize two prominent tensor decomposition methods, the Tucker decomposition and the Canonical Polyadic (CP) decomposition, to extract information contained in $\mathcal{S}$ and calculate the final uncertainty measures. CP decomposition represents the tensor as a sum of rank-one components, offering a more interpretable model. Its simplicity is advantageous in many scenarios, but it can be too rigid when the data exhibits rich interactions that a rank-one approximation fails to capture—potentially resulting in suboptimal solutions.  On the other hand, although Tucker decomposition excels at capturing high-order interactions, Tucker decomposition may converge to a local optimum due to its sensitivity to initialization and cause an unstable result. Therefore, in our paper, we use two tensor decomposition methods. We briefly introduce them below.

\noindent\textbf{Tucker Decomposition}. The Tucker decomposition is a higher-order generalization of the singular value decomposition (SVD) and represents a tensor as a core tensor multiplied by factor matrices along each dimension~\cite{de2000multilinear}. Formally, for a Nth-order tensor $\mathcal{X} \in \mathbb{R}^{I_1\times I_2 \times\cdots \times I_n}$,  the Tucker decomposition is expressed as:
\begin{equation}
     \mathcal{X} \approx \mathcal{G} \times_1 U^{(1)} \times_2 U^{(2)} \cdots \times_N U^{(N)},
\end{equation}
where \(\mathcal{G} \in \mathbb{R}^{R_1 \times R_2 \times \cdots \times R_N}\) is the core tensor that captures the interactions among the dimensions and \(U^{(i)} \in \mathbb{R}^{I_i \times R_i}\) for \(i=1,\dots,N\) are the factor matrices associated with each dimension.

Without loss of generality, we can assume that the Tucker factor matrices
$U^{(i)}$ are orthogonal and jointly form subspace bases of the different dimensions in the tensor. In that sense, Tucker decomposition jointly models the dimensions of the tensor while allowing the individual to model the subspaces of each dimension. Based on this feature, Tucker decomposition offers flexibility in choosing the ranks \((R_1, R_2, \dots, R_N)\) for different dimensions and each dimension \(n\) can have a different rank \(R_n\). 
A common way to compute the Tucker decomposition is via the Higher-Order Singular Value Decomposition (HOSVD)~\cite{de2000multilinear}.

\noindent\textbf{CP decomposition} Different from Tucker Decomposition, which might be sensitive to the noise in the tensor due to its high parameter complexity, CP decomposition offers a more robust alternative by representing the tensor as a sum of rank-one components. Specifically, for an \(N\)th-order tensor $\mathcal{X} $, the CP decomposition is expressed as:
\begin{equation}
    \mathcal{X} \approx \sum_{r=1}^{R} \lambda_r\, a^{(1)}_r \circ a^{(2)}_r \circ \cdots \circ a^{(N)}_r,
\end{equation}
where \(R\) is the CP rank, \(\lambda_r\) are scalar weights, \(a^{(n)}_r \in \mathbb{R}^{I_n}\) are the factor vectors associated with the \(n\)-th mode, and \(\circ\) denotes the outer product. CP decomposition can be computed using the Alternating Least Squares algorithm~\cite{kolda2009tensor}. Similar to the Tucker decomposition, the CP rank \(R\) can be chosen flexibly.

\noindent\textbf{Uncertainty} After we apply different tensor decomposition to $\mathcal{S}$, we can obtain the reconstructed tensor with different rank. For example, if we apply CP decomposition to $\mathcal{S}$ with rank $R$, we have:

\begin{equation}
    \mathcal{S} \approx \hat{\mathcal{S}}^{cp}_R = \sum_{r=1}^{R} \lambda_r\, a^{(1)}_r \circ a^{(2)}_r \circ \cdots \circ a^{(N)}_r
    \label{eq:cp_decomposition}
\end{equation}

Then we could compute the reconstruction error, which measures how much information could be captured with rank $R$:

\begin{equation}
    \epsilon^{cp}_R = \frac{\|\mathcal{S} - \hat{\mathcal{S}^{cp}_R}\|_F}{\|\mathcal{S}\|_F},
    \label{eq:cp_error}
\end{equation}
where \(\|\cdot\|_F\) is the Frobenius norm. Similarly, we could have a reconstruction tensor and error for Tucker decomposition with rank $[R,R,2]$:
\begin{equation}
    \epsilon^{tucker}_R = \frac{\|\mathcal{S} - \hat{\mathcal{S}^{tucker}_R}\|_F}{\|\mathcal{S}\|_F},
    \label{eq:tucker_error}
\end{equation}

Here, we always use the last mode uncompressed because we hope our method can analyze the two dimensions of responses separately. We also show an experiment that uses tucker with the last rank 1 in \cref{sec:ablation}.

\begin{figure}[ht]
    \centering
    \includegraphics[width=0.48\textwidth]{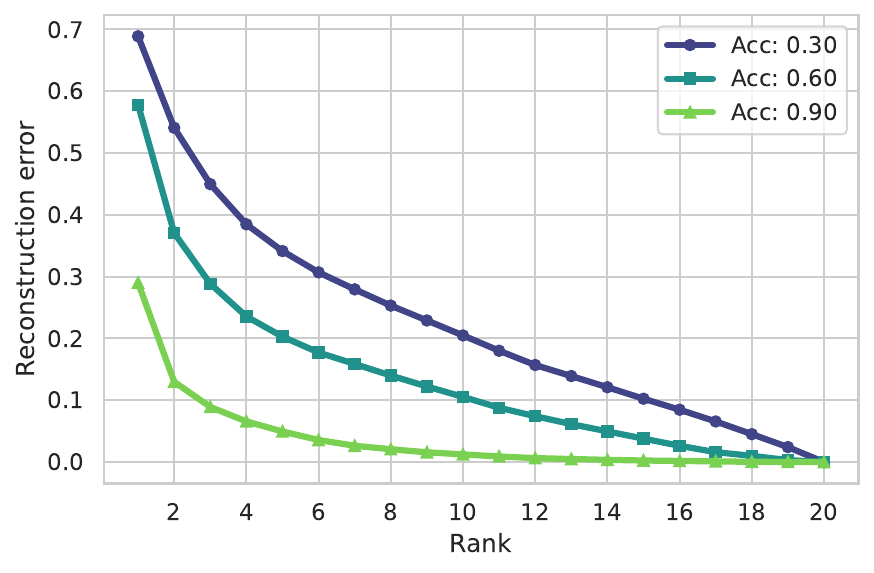}
    \vspace{-3mm}
    \caption{An plot of reconstruction errors v.s. rank for CP decomposition. The results show that a higher-accuracy sample tends to have a lower reconstruction error.}
    \label{fig:acc_vs_error}
\end{figure}

If the responses are more consistent, then $\mathcal{S}$ has an easier structure and thus it is easier to use a low-rank structure to capture the information. Therefore, the reconstruction error is expected to become lower when the responses are more consistent, i.e. the model is more confident. Therefore, we could directly use $\epsilon^{cp}_R$ or $\epsilon^{tucker}_R$ as the uncertainty measure.   To empirically prove this, we draw a figure to observe the relationship between different accuracy and reconstruction errors. We define the accuracy of a question to the mean accuracy of its $n$ responses. We present the results in \cref{fig:acc_vs_error}, which shows a higher accuracy sample indeed intends to have a lower reconstruction error.

\subsection{Ensemble Uncertainty}
\label{subsec:ensemble}

Though it is possible for us to directly use $\epsilon^{cp}_R$ or $\epsilon^{tucker}_R$ as the uncertainty measure, we further enhance the uncertainty measures by ensemble methods. Firstly, the choice of rank $R$ might influence the quality of $\epsilon^{cp}_R$ or $\epsilon^{tucker}_R$. Therefore, our method ensemble all the $\epsilon_R$ from $R=1$ to $R=n$. In detail, we have an uncertainty ensemble from all rank $R$ for CP decomposition:
\begin{equation}
    U_{cp} = \sum_{r=1}^n \epsilon^{cp}_R.
\end{equation}
Similarly, we compute the uncertainty ensemble from all rank $R$ for Tucker decomposition:
\begin{equation}
    U_{tucker} = \sum_{r=1}^n \epsilon^{tucker}_R.
\end{equation}

Finally, to ensure robust and accurate uncertainty estimation, we ensemble the uncertainty from both decomposition methods to obtain the final uncertainty. With two uncertainties from two tensor decomposition methods, we could have multiple meaningful ways to ensemble the results in an unsupervised way. We introduce two different methods with different intuitions. 

\noindent\textbf{Sum} A simple yet effective ensemble is the summation. The summation means that we assign an equal weight to the uncertainty results obtained from each rank and from both decomposition methods. This approach is based on the intuition that each decomposition method captures the different aspects of information and by adding two uncertainties up, we ensure that every method's contribution is treated evenly and thus obtain a more comprehensive information in the final uncertainty. In detail, we have:
\begin{equation}
    U = U_{\text{Tucker}} + U_{\text{CP}}.
\end{equation}

As mentioned above, a lower $U$ indicates a higher consistency between responses and thus a lower uncertainty. 

\noindent\textbf{Min}
The second ensemble strategy is based on the minimum operator. The intuition behind this approach is that if one of the tensor decomposition methods produces a low uncertainty estimate for a given sample, then the overall uncertainty should also be considered low. In other words, we rely on the most confident (i.e., smallest uncertainty) prediction provided by either method. Thus, we define the final uncertainty as:
\begin{equation}
    U = \min\left( U_{\text{Tucker}}, U_{\text{CP}} \right).
\end{equation}
This \textbf{Min} strategy can be particularly beneficial when one of the decomposition methods is prone to overestimating uncertainty. By selecting the minimum, we effectively mitigate the impact of any overly pessimistic estimates, while still preserving the robust information provided by the other method.

\begin{table*}[h]
    \centering
    \begin{tabular}{lcccccccccc}
        \toprule
        \multirow{2}{*}{Methods} & \multicolumn{2}{c}{Llama2-13B} & \multicolumn{2}{c}{Llama2-7B} & \multicolumn{2}{c}{Llama3.1-8B} & \multicolumn{2}{c}{Phi4} & \multicolumn{2}{c}{Deepseek-R1-7B} \\
        \cmidrule(lr){2-3} \cmidrule(lr){4-5} \cmidrule(lr){6-7} \cmidrule(lr){8-9} \cmidrule(lr){10-11}
        & AUROC & AUARC & AUROC & AUARC & AUROC & AUARC & AUROC & AUARC & AUROC & AUARC \\
        \midrule
        \multicolumn{11}{c}{\textbf{Dataset: CoQA} [Easy]} \\
        \midrule
        Eigv(Dis) &0.7294 & 0.7775&0.5965 & 0.9485 & 0.5762 & 0.9071 &0.6656 &0.9120 & 0.7841&0.9175 \\
        Ecc(Dis) & 0.6984& 0.7553& 0.5762 & 0.9409 & 0.5802 & 0.9206 &0.6487 &0.9066 & 0.7756& 0.9157\\
        Degree(Dis) & 0.7369 &0.7815 & 0.5963 & 0.9473 & 0.5728& 0.9112 & \textbf{0.6677} & \textbf{0.9121} & 0.7885&0.9189 \\
        Eigv(Agre) &0.7541 & 0.7876& 0.5971 & \textbf{0.9507} & 0.5791 &0.9153 &0.6399 &0.9051 & 0.7969&0.9234 \\
        Ecc(Agre) &0.7593 &0.7840 & 0.5961 &0.9480 &0.5785 & 0.9144 & 0.6335& 0.9020 & 0.7937& 0.9224\\
        Degree(Agre) &0.7548 & 0.7877& 0.5908&0.9413 & 0.5755& 0.9097&0.6278 & 0.8996& 0.7930 & 0.9222\\ \hline
        D-UE & 0.7566& \textbf{0.7885}& 0.5954&0.9481 &0.5825 &0.9284 &0.6503 &0.9079 &0.7966 &0.9228 \\
        P(true) & 0.7102 & 0.7088 & 0.5404 &0.9348 &0.5816 & \textbf{0.9323}&0.6630 &0.9087 & 0.5389&0.8311 \\ \hline
        \ours-Sum & \textbf{0.7657} &0.7739 & 0.5994 & 0.9483 & 0.5972  & 0.9308&0.6653 & 0.9076 & 0.7981 & 0.9253 \\
        \ours-Min &0.7656 &0.7740 & \textbf{0.5999} &0.9483 & \textbf{0.5983}& 0.9309 & 0.6652 & 0.9089& \textbf{0.7987} & \textbf{0.9255}\\ \hline
        \midrule
        \multicolumn{11}{c}{\textbf{Dataset: HotpotQA} [Medium]} \\
        \midrule
        Eigv(Dis) & 0.6269 & 0.7770& 0.6111& 0.7715 & 0.6099&0.6874 &0.5534 &0.8614 &0.5969 & 0.5737\\
        Ecc(Dis) &0.6103 & 0.7774& 0.6085&0.7752 &0.6044 &0.6827 & \textbf{0.5675} & 0.8691 & 0.5602 & 0.5377\\
        Degree(Dis) & \textbf{0.6336}& \textbf{0.7790} & 0.6134 &0.7714 & 0.6202&0.7087 &0.5666 & 0.8590 & 0.5977&0.5756 \\
        Eigv(Agre) &0.6235 & 0.7638 &0.6035 &0.7648 &0.6176 &0.7035 & 0.5328& 0.8497& 0.6249 & 0.5878\\
        Ecc(Agre) & 0.6233&0.7670 & 0.6049&0.7666 & 0.6084&0.6991 &0.5469 &0.8594&0.6321 & 0.5907\\
        Degree(Agre) & 0.6217 & 0.7611& 0.5973 & 0.7600 & 0.6105&0.7016 & 0.5294& 0.8491&0.6278 & 0.5902\\ \hline
        D-UE & 0.6252&0.7659 & 0.6056&0.7669 & 0.6212&0.7083 &0.5335 &0.8511 & 0.6270 &0.5893 \\
        P(true) &0.6056 &0.7591 &0.5901 &0.7713 & 0.6362&0.7077 &0.5326 &0.8513 & 0.5081 & 0.4795\\ \hline
        \ours-Sum  &0.6292 & 0.7730 & \textbf{0.6163}& 0.7770& \textbf{0.6471}& \textbf{0.7235} & 0.5669 &0.8735 & \textbf{0.6331} & \textbf{0.5918} \\
        \ours-Min  & 0.6296 & 0.7735& 0.6155&\textbf{0.7771}& \textbf{0.6471}&0.7234 &0.5673 & \textbf{0.8736} & 0.6329 & 0.5912 \\
        \hline
        \midrule
        \multicolumn{11}{c}{\textbf{Dataset: NQ\_Open} [Hard]} \\
        \midrule
        Eigv(Dis) & 0.6162&0.7300 &0.7280 & 0.6367& 0.6742& 0.5343&0.7035 &0.6035 &0.6696 & 0.2451\\
        Ecc(Dis) &0.6210 &0.7330 &0.7167 &0.6172 & 0.6562& 0.5007&0.6898 &0.5828 &0.6607 &0.2237 \\
        Degree(Dis) &0.6130 &0.7168 &0.7273 & 0.6318&0.6865 &0.5430 & 0.7090&0.6094 &0.6675 & 0.2463\\
        Eigv(Agre) &0.6258 &0.7276 &0.7240 &0.6327 &0.7463 & 0.5801&0.7515 & 0.6351 & 0.7291&0.2758 \\
        Ecc(Agre) & 0.6273 &0.7311 &0.7307 & 0.6298&0.7612 &0.5875 & 0.7555& 0.6370 & 0.7437& 0.2836\\
        Degree(Agre) & 0.6286 &0.7355 & 0.7290& 0.6324& 0.7619 &0.5885 &0.7542 & 0.6341 &0.7353 & 0.2781\\ \hline
        D-UE &0.6281 &0.7320 & 0.7258&0.6342 &0.7551 &0.5833 & 0.7539 &0.6366 & 0.7333 & 0.2801 \\
        P(true) & 0.6197& 0.7289& 0.6532&0.5929 & 0.7061& 0.5522& 0.7096&0.6049 & 0.4865&0.1413 \\ \hline
        \ours-Sum & \textbf{0.6334} & \textbf{0.7410} & 0.7310& 0.6359 & \textbf{0.7642} & \textbf{0.5906} & 0.7562 & 0.6381 & \textbf{0.7560} & \textbf{0.2843} \\
        \ours-Min &0.6332 &0.7409 & \textbf{0.7313} & \textbf{0.6375} & 0.7641 & 0.5905& \textbf{0.7565} & \textbf{0.6386} & 0.7558 & 0.2842 \\ \hline
        \bottomrule
    \end{tabular}
    \caption{Comparison of our methods with different baselines on various datasets and large language models. The best result is shown in the \textbf{bold}. The results show that \ours performs better than baselines in general and \ours has a better advantage on more difficult datasets such as NQ\_Open.}
    \vspace{-5mm}
    \label{tab:main_results}
\end{table*}

\section{Experiments}

We conduct comprehensive experiments across multiple datasets and model architectures to validate our method's ability to decouple explanation robustness from classification robustness. Our evaluation addresses three key research questions:
\begin{itemize}
    \item \textbf{RQ1:} Does \ours have better quantify uncertainties?
    \item \textbf{RQ2:} How do different ensemble methods and information from both dimensions help?
    \item \textbf{RQ3:} Is\ours robust to different settings? 
\end{itemize}

\begin{table}[H]
\centering
\resizebox{!}{0.11\textwidth}{
\begin{tabular}{@{}lc@{}}
\toprule
\textbf{Measure} & \textbf{Details} \\ 
\midrule
$U_{\textit{Eigv}}(Dis)$ & \multicolumn{1}{c}{Spectral eigenvalue on the disagreement.} \\ 
$U_{\textit{Ecc}}(Dis)$ & \multicolumn{1}{c}{Average distance in responses' disagreement.} \\ 
$U_{\textit{Degree}}(Dis)$ & \multicolumn{1}{c}{Degree of disagreement similarity Matrix.} \\ 
$U_{\textit{Eigv}}(Agre)$ & \multicolumn{1}{c}{Spectral eigenvalue on the agreement.} \\ 
$U_{\textit{Ecc}}(Agre)$ & \multicolumn{1}{c}{Average distance in responses' agreement.} \\ 
$U_{\textit{Degree}}(Agre)$ & \multicolumn{1}{c}{Degree Matrix of agreement Matrix.} \\ 
$p(true)$ & \multicolumn{1}{c}{Entropy of knowledge dimension responses} \\ 
$D-UE$ & \multicolumn{1}{c}{eigenvalue from Laplacian of a directional graph} \\ 

\bottomrule
\end{tabular}}
\vspace{-1mm}
\caption{The baseline methods and explanations.}
\vspace{-5mm}
\label{tab:baslines}
\end{table}

\subsection{Experimental Setup}
\label{sec:setup}
\subsubsection{Datasets} As mentioned in \cref{sec:background}, following prior works~\cite{lin2022teaching}, we focus on open-form question-answering 
(QA) tasks in this paper. We adopt 4 different classic QA datasets. Coqa~\cite{reddy2019coqa} is a conversational question-answering dataset that contains dialogues with free-form answers grounded in diverse passages, which is the easiest dataset among all datasets. HotpotQA~\cite{yang2018hotpotqa} is a multi-hop QA dataset that demands reasoning over multiple Wikipedia paragraphs to derive correct answers. NQ-Open~\cite{kwiatkowski2019natural} consists of real-world queries from Google Search, requiring models to retrieve and answer questions without explicit context, which is the hardest dataset. 
\subsubsection{Models to Evaluate} We evaluate \ours on Llama family~\cite{touvron2023llama}, which is the one of the most popular LLMs. In detail, we use Llama-2-13b and Llama-2-7B to demonstrate the effectiveness of \ours with different model sizes and use Llama-3.1-8B~\cite{dubey2024llama} to that \ours could also work on the different version of Llama. To further demonstrate the generalization ability for other architectures,  we also use Phi4~\cite{abdin2024phi} and Deepseek-R1-distill-7B~\cite{guo2025deepseek} in our paper.


\subsubsection{Evaluation Metrics} Effective uncertainty measures should accurately represent the reliability of LLM responses, with higher uncertainty more likely leading to incorrect generations and vice versa~\cite{lin2023generating,kuhn2023semantic}. Following prior works~\cite{lin2023generating,da2024llm}, we mainly use UQ values to predict whether an answer is correct or not. Following prior works~\cite{lin2023generating,da2024llm}, we will use Area Under Receiver Operating Characteristic (AUROC) and Area Under Accuracy Rejection Curve (AUARC) as evaluation metrics, where a higher AUROC or AUARC demonstrates better uncertainty measures. To compute AUROC and AUARC, the accuracy of each original response is required. Following previous works~\cite{da2024llm,lin2023generating}, we use another LLM to provide correctness from 0-100 to each response. If the correctness is greater than 70, we label the response as correct. In this paper, we use Qwen-34B~\cite{bai2023qwen} to evaluate the correctness.

\subsubsection{Knowledge Extracted Models} In this paper, we mainly use llama-2-13b~\cite{touvron2023llama} as the auxiliary models to extract the knowledge dimension of responses. To demonstrate the robustness of \ours with different knowledge-extracted models, we also contain the results for different LLMs as knowledge-extracted models.

\subsubsection{Baselines} In this paper, we compared \ours with baselines that use semantic dimension response and knowledge dimension response. For semantic dimension, we mainly compared with methods that come from \citet{lin2023generating}. In detail, we incorporate six distinct methods from \citet{lin2023generating}, which differ based on the operations applied after computing similarity and whether they utilize agreement (entailment) probabilities or disagreement (contradiction) logits to construct the similarity matrix. For knowledge dimension, we use D-UE~\cite{da2024llm} and $p(true)$~\cite{kadavath2022language} as the baselines. Note that we use $p(true)$ on the knowledge dimension of response. We show the detailed explanations of all baselines in \cref{tab:baslines}

\begin{figure*}[t]
\centering
\begin{minipage}[t]{0.32\linewidth}
  \centering
  \includegraphics[width=\linewidth,trim=0 0 0 1cm, clip]{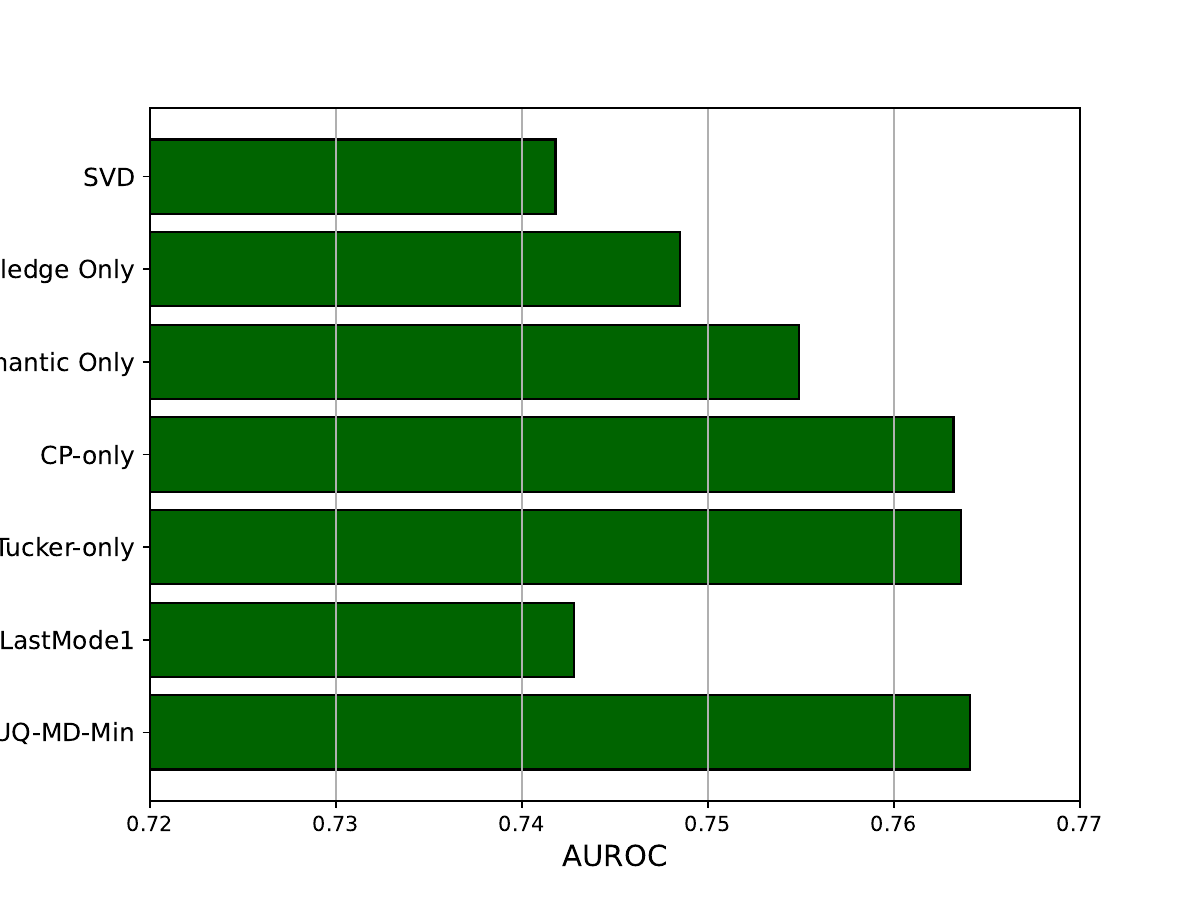}
  \captionof{figure}{Ablation studies that show that \ours fully utilizes all the information from both dimensions.}
  \label{fig:ablation}
\end{minipage}\hfill
\begin{minipage}[t]{0.32\linewidth}
  \centering
  \includegraphics[width=\linewidth]{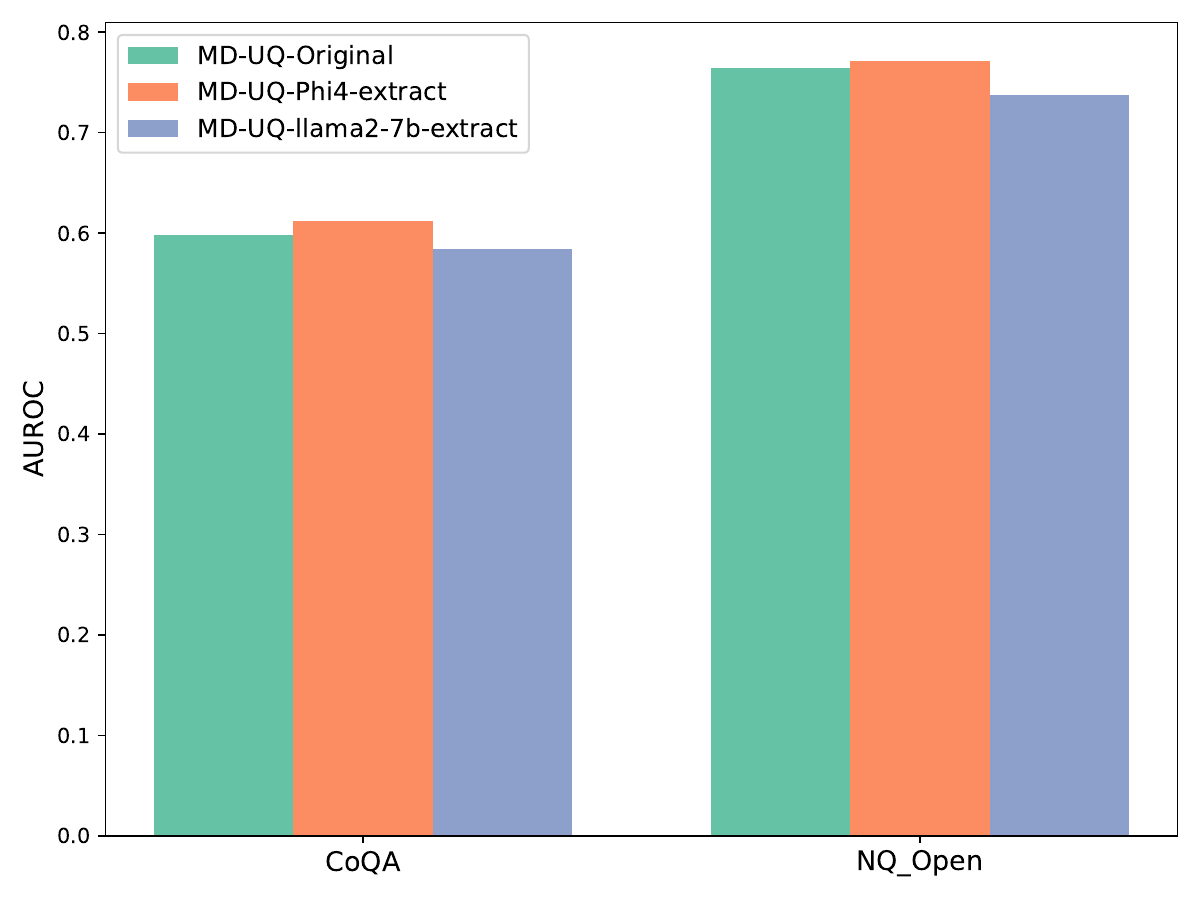}
  \captionof{figure}{Performance for different knowledge extract models on CoQA and NQ\_Open with llama3.1.}
  \label{fig:knowledge_extract}
\end{minipage}\hfill
\begin{minipage}[t]{0.32\linewidth}
  \centering
  \includegraphics[width=\linewidth]{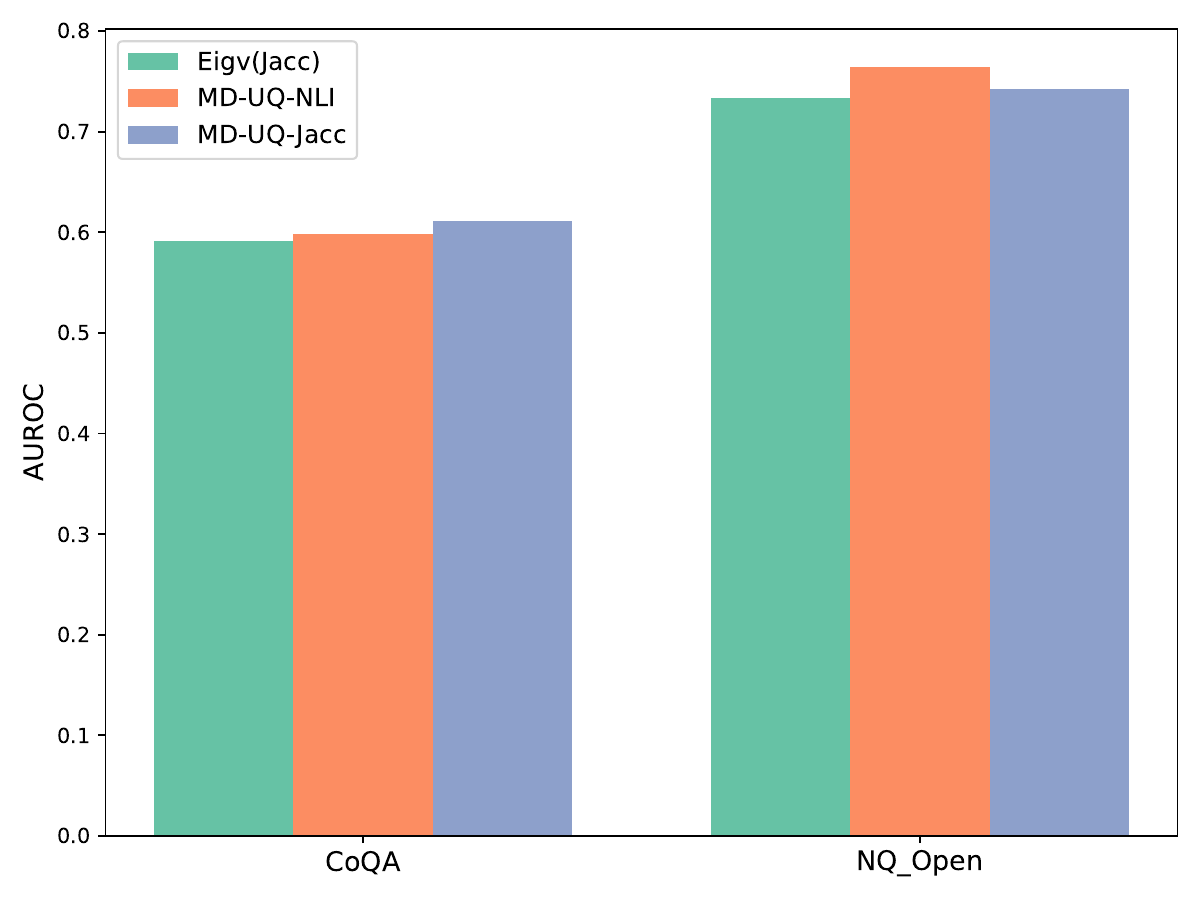}
  \captionof{figure}{Performance that uses Jaccard similarity on CoQA and NQ\_Open with llama3.1.}
  \label{fig:jacc}
\end{minipage}
\end{figure*}

\subsection{Does \ours have better quantify uncertainties? (RQ1)}
\label{sec:main_result}
In this section, we explore whether \ours has better uncertainties compared with state-of-the-art uncertainty quantification methods. In \cref{tab:main_results}, we compare \ours with 8 baselines across three different datasets and five different models as introduced in \cref{sec:setup} In detail, we have the following observations:

\noindent $\bullet$ Compared with all baseline methods, \ours achieves the best performance overall. Especially when we consider AUROC. For AUARC, \ours achieves the best performance for NQ\_Open while \ours also achieves the comparable performance for CoQA in most scenarios. These results demonstrate that \ours has better quantify uncertainties overall. \\
\noindent $\bullet$ Among all datasets, \ours achieves the highest performance improvement on NQ\_Open, which is the most difficult dataset among all datasets and may lose to baselines for an easier dataset like CoQA. This indicates \ours could perform even better when the task is harder, where uncertainty quantification is more important. \\
\noindent $\bullet$ Two different ensemble methods show very similar results. Min strategy performs better than the sum strategy under $61.51\%$ situations, indicating that difficult datasets might also have more complex structures that single one tensor decomposition might oversight some information while using min structure could reduce such oversight by considering the best cases. However, both ensemble methods show a better performance than all baselines, which proves the effectiveness of tensor decomposition. \\

From these results, we get a conclusion that overall, \ours have better uncertainties.

\subsection{How do different ensemble methods and information from both dimensions help? (RQ2)}
\label{sec:ablation}
In this section, we use more experiments to prove the necessity of using information from both semantic and knowledge dimensions as well as using tensor decomposition. In detail, we consider the following methods: 1) \ours with only semantic responses, 2) \ours with only knowledge responses and 3) Concatenating similarity matrices from semantic and knowledge dimensions into a 2D matrix and applying SVD, 4) only using one tensor decomposition. In \cref{fig:ablation}, we show the comparison between \ours and other methods.  The results show that \ours consistently outperforms its variants and SVD method that repeated information will dominate the features, showing the effectiveness of our framework.

\subsection{Is \ours robust to different settings? (RQ3)}
\subsubsection{Different Knowledge Extracted Models} Knowledge extracted models influence the claim extraction in \ours as stated in \cref{subsubsec:knowledge}. Therefore, in this section, We test the robustness of \ours on various knowledge extracted models. unlike using llama2-13b in \cref{sec:main_result} and \cref{sec:ablation}, we conduct experiments on CoQA and NQ\_open using llama2-7b and llama3.1 as the knowledge extracted models, We show the results in \cref{fig:knowledge_extract}. From the figure, we can see that using Phi4 could even achieve a better result, indicating \ours has more potential with the development of LLMs. 

\subsubsection{Different Accuracy Thresholds} Different accuracy thresholds lead to different accuracy and influence the evaluation of uncertainties. In the previous experiments, we all set the accuracy threshold to 70 as mentioned in \cref{sec:setup}.  To test the robustness of \ours under different accuracy thresholds, we choose an extra dataset TriviaQA~\cite{joshi2017triviaqa}, which is considered the easiest dataset, and NQ\_Open, which is the most challenging dataset in our paper to conduct experiments. We show the results with accuracy thresholds of 70 and 90 in \cref{tab:Accuracy_threshold}. From the results, we can see that increasing the accuracy threshold decreases the performance of all baselines while the performance of \ours could even increase for datasets with different difficulties, showing the robustness of \ours in different settings. 

\subsubsection{Different Similarity Metrics} Finally, different similarity metrics lead to different similarity matrices. Therefore, to test whether \ours also has a good performance for different similarities, we use Jaccard similarity instead of using an NLI model in this section and the results are presented in \cref{fig:jacc}. The results show that using Jaccard similarity will boost the performance for a simple dataset like CoQA but hurt the performance for a difficult dataset like NQ\_Open. This is because the answer to a simple question might not have a deeper semantic meaning that requires NLI models. However, \ours can still outperform baseline methods that also use Jaccard similarity, showing the robustness of \ours.

\begin{table}[h]
    \centering
    \resizebox{0.5\textwidth}{!}{
    \begin{tabular}{lcccc}
        \toprule
        \multirow{2}{*}{Methods} & \multicolumn{2}{c}{Accuracy Threshold: 0.7} & \multicolumn{2}{c}{Accuracy Threshold: 0.9} \\
        \cmidrule(lr){2-3} \cmidrule(lr){4-5}
        & AUROC & AUARC & AUROC & AUARC \\
        \midrule
        \multicolumn{5}{c}{\textbf{Dataset: TriviaQA} [Easy]} \\
        \midrule
        Eigv(Dis) & 0.8261 & 0.8094 & 0.8100& 0.7604\\
        Ecc(Dis) & 0.8063& 0.7940&0.7892 & 0.7415\\
        Degree(Dis) &0.8399 & 0.8163&0.8259 & 0.7694\\
        Eigv(Agre) &0.8436 &0.8116 &0.8351 & 0.7721 \\
        Ecc(Agre) & \textbf{0.8510}&0.8189 & 0.8374&0.7721 \\
        Degree(Agre) &0.8396 &\textbf{0.8397} &0.8384 & 0.7739\\
        \ours-Sum &0.8428 &0.8144 & 0.8438&0.7749 \\
        \ours-Min &0.8431 &0.8149 & \textbf{0.8440} & \textbf{0.7754}\\
        \midrule
        \multicolumn{5}{c}{\textbf{Dataset: NQ\_Open} [Hard]} \\
        \midrule
        Eigv(Dis) & 0.6162 & 0.7300 &0.5636 &0.6017 \\
        Ecc(Dis) & 0.6210& 0.7330& 0.5658&0.5941 \\
        Degree(Dis) &0.6130 & 0.7168&0.5662 &0.6033 \\
        Eigv(Agre) &0.6258 &0.7276 & 0.6146& 0.6290 \\
        Ecc(Agre) & 0.6273&0.7311 &0.6239 &0.6344\\
        Degree(Agre) &0.6286 &0.7355 & 0.6221&0.6299 \\
        \ours-Sum &\textbf{0.6334} &\textbf{0.7410} &\textbf{0.6351} &\textbf{0.6430} \\
        \ours-Min &0.6332 &0.7409 & 0.6350 &0.6429 \\
        \bottomrule
    \end{tabular}
    }
    \caption{Comparison of different methods across different accuracy thresholds on TrivialQA and NQ\_Open with llama2-13B. The results show that our methods outperform baselines after increasing the accuracy threshold, indicating that our methods have an advantage on more difficult datasets.}
    \vspace{-7mm}
    \label{tab:Accuracy_threshold}
\end{table}

\section{Related Works}

Uncertainty quantification for traditional machine learning problems such as regression or classification has been well studied~\cite{ye2024uncertainty,amini2020deep,sensoy2018evidential,ovadia2019can}. Most previous works on uncertainty quantification for nature language processing (NLP) consider text classifiers~\cite{jiang2021can,desai2020calibration,kamath2020selective} or text regressors~\cite{glushkova2021uncertainty,wang2022uncertainty}. To transfer NLP tasks into a classification task, previous work may consider using multi-choice question answering datasets or transferring questions into multi-choice form~\cite{kamath2020selective}. 

However, considering NLP tasks as simple classification tasks overlooks the generation nature of the LLMs~\cite{kuhn2023semantic}. To overcome this disadvantage, recent works focus on open-ended generation tasks. The first branch of research is inducing the LLMs to output their uncertainty along with the response to solve UQ for open-ended generation tasks~\cite{tian2023just,kadavath2022language,mielke2020linguistic}. \citet{lin2022teaching} even fine-tuned LLMs so that LLMs can output better uncertainty. This is a straightforward solution. However, fine-tuning the LLMs to obtain a better uncertainty measure requires white-box access to the models and may cost computation resources. \citet{kuhn2023semantic} first propose semantic entropy, which calculates entropy considering semantic information. However, such an approach still requires the token-related probability values as input.

To compute uncertainty for black-box MLLMs, previous works take a step further compared with semantic entropy and utilize the similarity and consistency between different generated answers from the same query to the LLMs. \citet{lin2023generating} uses NLI models to obtain the similarity. Then they treat the similarity matrix as from a weight connected graph and compute uncertainty using the structure of the graph such as using eigenvalues from the graph Laplacian. \citet{chen2024quantifying} identify unreliable or speculative answers by computing a confidence score. However, both works only consider semantic similarity, lacking an analysis of the deep meaning of the output. \citet{da2024llm} contains a claim level response augmentation. However,  augmented responses share much common information with original responses, and such common information is not considered by \citet{da2024llm}, causing a potential performance downgrade.  Therefore, in our paper, we do not only generate implicit knowledge behind the original answers but also use tensor decomposition to fully utilize the additional information.  

\vspace{-3mm}
\section{Conclusion}
In conclusion, this study introduces a novel multi-dimensional uncertainty quantification framework, MD-UQ, for large language models, addressing the limitations of conventional uncertainty estimation approaches. By leveraging both semantic similarity and knowledge coherence dimensions, our method disentangles and integrates complementary information to achieve a more robust uncertainty representation. Through the application of tensor decomposition techniques, MD-UQ effectively reduces redundant information and enhances the reliability of uncertainty assessments. Experimental results across multiple datasets and models demonstrate the superiority of our framework in distinguishing uncertain responses, particularly in complex and high-stakes environments.

\bibliographystyle{ACM-Reference-Format}
\bibliography{sample-base}
\end{document}